\documentclass[conference]{IEEEtran}
\IEEEoverridecommandlockouts
\usepackage{float}

\usepackage{amsmath,amssymb,amsfonts}
\usepackage[ruled,vlined]{algorithm2e}
\usepackage{algorithmic}
\usepackage{commath}
\usepackage{amsmath}
\usepackage{graphicx}
\usepackage{tabto}
\usepackage{textcomp}
\usepackage{xcolor}
\usepackage{hyperref}
\usepackage{cite}
\usepackage{multirow}
\usepackage{lipsum}
\usepackage{mwe}
\usepackage{comment}
\usepackage[font=smaller]{caption}
\usepackage[T1]{fontenc}
\usepackage[super]{nth}
\usepackage{textcomp}
\usepackage{color, colortbl}
\definecolor{LightGray}{gray}{0.85}
\definecolor{LightCyan}{rgb}{0.88,1,1}
\usepackage{xcolor}
\usepackage[export]{adjustbox}
\usepackage{enumitem}
\setlist[itemize]{noitemsep, nolistsep}
\usepackage{titlesec}
\def\BibTeX{{\rm B\kern-.05em{\sc i\kern-.025em b}\kern-.08em
    T\kern-.1667em\lower.7ex\hbox{E}\kern-.125emX}}

\begin{document}

\title{\LARGE \bf Variable Grasp Pose and Commitment for Trajectory Optimization}

\author{Jiahe Pan$^{1}$, Kerry He$^{2}$, Jia Ming Ong$^{2}$, Akansel Cosgun$^{3}$
\thanks{\hspace*{-1em}$^{1}$University of Melbourne, Australia\newline $^{2}$Monash University, Australia\newline $^{3}$Deakin University, Australia}}

\maketitle

\begin{abstract}

We propose enhancing trajectory optimization methods through the incorporation of two key ideas: variable-grasp pose sampling and trajectory commitment. Our iterative approach samples multiple grasp poses, increasing the likelihood of finding a solution while gradually narrowing the optimization horizon towards the goal region for improved computational efficiency. We conduct experiments comparing our approach with sampling-based planning and fixed-goal optimization. In simulated experiments featuring 4 different task scenes, our approach consistently outperforms baselines by generating lower-cost trajectories and achieving higher success rates in challenging constrained and cluttered environments, at the trade-off of longer computation times. Real-world experiments further validate the superiority of our approach in generating lower-cost trajectories and exhibiting enhanced robustness. While we acknowledge the limitations of our experimental design, our proposed approach holds significant potential for enhancing trajectory optimization methods and offers a promising solution for achieving consistent and reliable robotic manipulation.

\end{abstract}


\section{Introduction} \label{section:intro}

Robotic manipulation techniques have traditionally followed a two-step process \cite{newbury2023deep}. Firstly, a grasp pose is determined based on the likelihood of grasp success using either model-based or deep learning-based approaches. Subsequently, the arm trajectory is computed by employing sampling-based planning methods such as Rapidly-exploring Random Trees (RRT) \cite{RRTconnect} or Probabilistic Roadmaps (PRM) \cite{prm}, or through trajectory optimization algorithms such as Covariant Hamiltonian Optimization for Motion Planning (CHOMP) \cite{chomp} or Sequential Convex Optimization (TrajOpt) \cite{schulman2014motion}. However, solely selecting the end-effector goal pose based on predicted grasp success without considering the arm motion can lead to suboptimal arm trajectories in terms of factors like completion time or energy efficiency. This limitation presents an opportunity for improvement, particularly when grasp poses have similar probabilities of success or when the robot operates in cluttered environments where different grasp poses may yield vastly different trajectories.

Previous works have explored various approaches for trajectory planning with variable grasp poses. Dragan et al. \cite{srinivasa2011manipulation} extended CHOMP \cite{chomp} to include goal sets as end-point constraints. Murooka et al. \cite{murooka} combined sampling-based planning and gradient-based optimization to solve for optimal trajectories that reach towards sampled grasp poses. Horowitz et al. \cite{horowitz} explored grasp and manipulation planning as a single optimal control problem. Ichnowski et al. \cite{gomp} used Dex-Net 4.0 \cite{dexnet4.0} and an SQP-based algorithm for optimizing minimum-time trajectories in warehouse picking tasks.

\begin{figure}[ht]
\centerline{\includegraphics[trim={11 6 4 6}, clip, scale=0.52]{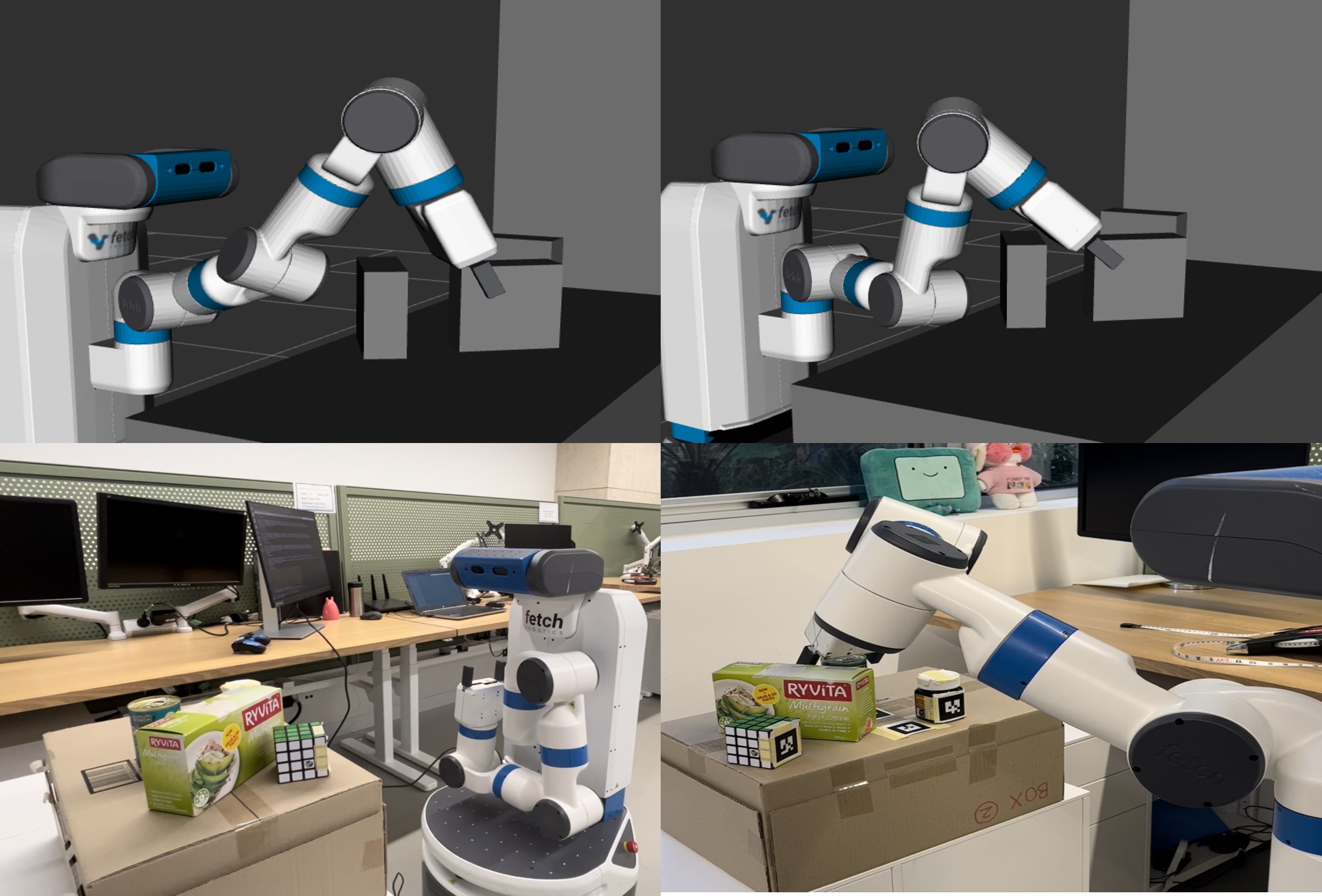}}
\caption{Trajectory planning for various grasp poses for an occluded cuboid target (top). The robot arm starts in a highly constrained position (arm-tuck), and plans a trajectory towards grasping a cylindrical target (bottom).}
\label{fig:intro_pic}
\end{figure}

Vahrenkamp et al. \cite{integrated} evaluated trajectories towards different grasp poses using an online quality measurement module. Jetchev et al. \cite{jetchev2010trajectory} used 3D laser point cloud information to construct an occupancy probability map for trajectory generation in cluttered environments. Dogar et al. \cite{dogar2012physics} analyzed physical contacts with cluttered scenes to improve planner efficiency and success rates. Berenson et al. \cite{wgr} used goal regions in the manipulator's workspace for grasp and trajectory planning. Bergman et al. \cite{bergman2020optimization} proposed Receding Horizon Planning (RHP), and Mandalika et al. \cite{mandalika2018lazy} introduced Lazy Receding Horizon A* for path planning in cluttered scenes.

In this paper, we propose an approach that incorporates the selection of grasp poses into trajectory optimization. The ability to vary the grasp pose relative to the target object becomes particularly advantageous in cluttered environments. For example, daily household objects can be grasped in multiple ways, and the chosen grasp pose directly affects the quality and feasibility of the robot arm's trajectory during the reach and grasp actions, especially in constrained task environments. Therefore, our approach aims to explore the space of all possible grasp poses for a given target object and select one that leads to a more efficient trajectory for performing the grasping action. Additionally, we explore trajectory commitment in this paper in which we gradually shrink the optimization horizon towards the goal region to increase the efficiency of trajectory optimization. To validate our approach, we implement several baseline methods based on TrajOpt and RRT-Connect. We compare these baselines against our approach in four distinct task environments featuring primitive targets and obstacles arranged in different configurations, with varying initial arm poses. Finally, we evaluate our method by analyzing planning time, trajectory cost, and success rates of the different approaches in simulated and real-world experiments on a 7-DoF serial manipulator.

Our approach differs from existing methods by directly sampling grasp poses during planning, without pre-computing or ruling out infeasible grasps. We treat TrajOpt as a black-box optimizer, refining segments of the trajectory near the target object. Our refinement step ensures optimality of the entire trajectory. Notably, our method is accessible to researchers without extensive knowledge of grasp-pose synthesis or trajectory planning with high-DOF robotic manipulators, and despite treating TrajOpt as a black-box optimizer, users can still control its high-level input parameters which directly relate to the trajectories generated.


\section{Proposed Approach} \label{section:proposed_approach}

We build upon the original TrajOpt algorithm \cite{schulman2014motion}, treating it as a black-box optimizer. Given information about the task environment, an initial robot arm pose, and a goal pose, TrajOpt outputs the optimal trajectory from the initial to the final goal pose. We augment TrajOpt with an outer optimization loop that randomly samples different grasp poses relative to the target object. For each sampled pose, we use TrajOpt to solve for an optimal trajectory.

A key hypothesis behind our commitment idea is that since only the grasp pose varies between iterations, the waypoints near the target object are more affected than those near the starting pose. Therefore, it is unnecessary to recompute the entire trajectory in each iteration. Instead, we gradually shrink the optimization horizon towards the goal region to increase TrajOpt's efficiency. This reduction in the optimization horizon helps the manipulator escape from constrained configurations and navigate clustered environments. We define the waypoints outside TrajOpt's optimization horizon as committed waypoints, which are stored in memory. As the outer-loop iterations progress, we add new committed waypoints and further shrink the optimization horizon towards the goal region. 

Sec. \ref{sec:black_box} provides a brief overview of the TrajOpt algorithm and illustrates how it is embedded as a black-box optimizer in our algorithm. Sec. \ref{sec:grasp_pose_parametrization} presents how we parameterize the variable grasp pose for a primitive object shape. In Sec 
\ref{sec:algorithm}, we provide a description of our algorithm and illustrate the high-level logic using a flowchart.

\subsection{Black-box Optimizer}
\label{sec:black_box}

Trajectory optimization for robotic manipulation is often posed as a non-convex, constrained optimization problem. In the TrajOpt implementation, the cost function is defined as the squared sum of the accumulative velocity, acceleration and jerk costs across consecutive joint-space waypoints of the trajectory, which relates to the length of the physical trajectory. Equality and inequality constraints arise from the robot's joint limits, avoiding collision objects, and reaching the goal pose. The optimization problem is solved iteratively through a series of convex optimization problems using SQP. The penalty coefficients for the constraints vary according to improvement in the cost, until it converges and a valid solution trajectory is found, or an iteration limit has been reached. 

Our algorithm interacts directly with the following 5 parameters of the TrajOpt algorithm:
\begin{itemize}

    \item \textbf{Start pose} - The initial state of the robot.

    \item \textbf{Goal pose} - The target pose that TrajOpt will plan the trajectory towards.

    \item \textbf{InitTraj} - The initial trajectory given to TrajOpt to initialize the SQP algorithm.

    \item \textbf{Tolerance} - The tolerance for constraint-violation. Values greater than this parameter leads to a trajectory being considered invalid by TrajOpt.

    \item \textbf{Waypoints} - The number of joint-space waypoints to plan for given some initial and final poses.
    
\end{itemize}
Overall, the TrajOpt algorithm is treated as a black-box optimizer which takes inputs of planning queries and outputs feasible solution trajectories if found.

\subsection{Grasp Pose Parametrization} \label{sec:grasp_pose_parametrization}

\begin{figure}[ht]
\centerline{\includegraphics[trim={11 6 4 6}, scale=0.38]{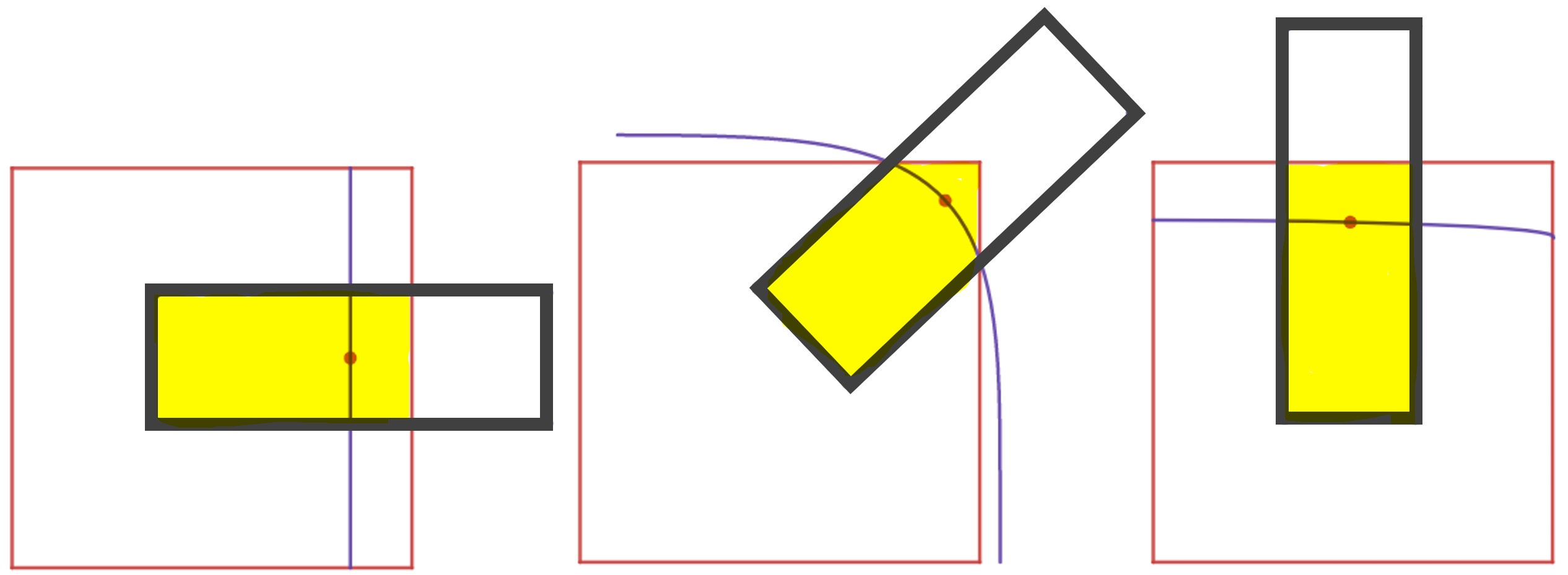}}
\caption{Side view of grasp angles of \(0^{\circ}, 45^{\circ}\) and \(90^{\circ}\) (left to right). Red dot is the center point position of the end-effector, purple line shows the possible positions of \(s\), overlap regions are highlighted.} 
\label{fig:grasp_2d}
\end{figure}

To introduce how we paramterize the set of possible ways a robot can grasp a primitive object, let us first consider a cuboid. When the robot faces this object directly, it has 3 feasible grasp direction - \{front, left, right\}, and the extra grasp pose from the top face and all the grasp angles in between. First analyzing a single side (see Fig. \ref{fig:grasp_2d}), we define the parameters \((\theta, s)\), where \(\theta\) governs the grasp angle from horizontal, and \(s\) is the location of the end-effector's centre point along an arc which varies with \(\theta\). This parametrization model maintains a high overlap-region between the cross-sections of the end-effector and the object, hence ensuring stable grasps for all feasible poses. The bounds for \((\theta, s)\) are set relative to the object's dimensions, allowing our model to scale automatically to different-sized objects.

\vspace{-0.2cm}

\begin{figure}[ht]
\centerline{\includegraphics[trim={0 20 0 22}, clip, scale=0.4]{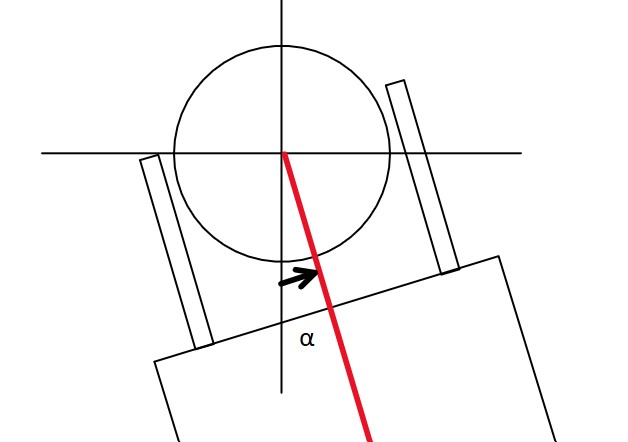}}
\caption{(Top view) End-effector grasping a round object. The parameter \(\alpha\) introduces an extra degree of freedom in the grasp pose representing the approach angle.}
\label{fig:3d_grasp}
\end{figure}

Our model can easily generalize to cylindrical and spherical objects by adding a rotational degree of freedom in the parametrization, using the parameter $ \alpha $ (see Fig. \ref{fig:3d_grasp}).

\subsection{The Algorithm}  \label{sec:algorithm}

\renewcommand{\arraystretch}{1.3}

\begin{table}[t]
  \begin{center}
    \begin{tabular}{|c|p{7.3cm}|}
    \hline
    \textbf{Symbol} & \textbf{Description} \\ \hline
      $ \mathbf{\mu_{max}} $ & A predefined tolerance threshold ensuring that the final solution trajectory satisfies goal constraints and is collision-free. \\ \hline
      $ \mu_0 $ & The initial value of the penalty coefficient, initialized at two orders of magnitude higher than $\mathbf{\mu_{max}}$. This relaxes the problem initially, aiding TrajOpt in generating trajectories efficiently, especially for highly constrained environments.\\ \hline
      $ \mathrm{MaxIters} $ & The iteration limit for the outer optimization loop, independent of any iteration limits within the TrajOpt algorithm. \\ \hline
      $ \mathbf{q}_s $ & Initial robot configuration from the problem description \\ \hline
      $ N_0 $ & 	A fixed parameter defining the total number of waypoints to plan for over the entire horizon. \\ \hline
      $ \mathbf{q}_0 $ & The current starting robot configuration (\textbf{Start pose}) \\ \hline
      $ \mathbf{G} $ & The randomly-sampled Cartesian grasp pose in each iteration. (\textbf{Goal pose}) \\ \hline
      $ \mathbf{x}_0 $ & The linearly-interpolated trajectory between $\mathbf{q}_0$ and $\mathbf{G}$, unless in the refinement step, where TrajOpt is initialized with the concatenated trajectory. (\textbf{InitTraj}) \\ \hline
      $ \mu $ & The current constraint-tolerance (\textbf{Tolerance})  \\ \hline
      $ N $ & The current number of waypoints inside the optimization horizon, initially set to $N_0$. (\textbf{Waypoints}) \\ \hline
      $ \mathcal{C} $ & The list of committed trajectory waypoints, updated in each commitment step. \\ \hline
      $ s_{thres} $ & A threshold variable that determines when commitment steps are triggered. Initialized as 1 and incremented at each commitment step, encouraging exploration before making further commitments. \\ \hline
      $ s $ & The count of successful TrajOpt runs for the current optimization horizon. Resets to 0 after each commitment step.\\ \hline
      \end{tabular}
      \caption{Summary of parameters and variables. Highlighted variables are used as inputs to the TrajOpt algorithm.}
      \label{tab:params_variables}
  \end{center}
\end{table}

\begin{figure}
\centerline{\includegraphics[scale=0.8]{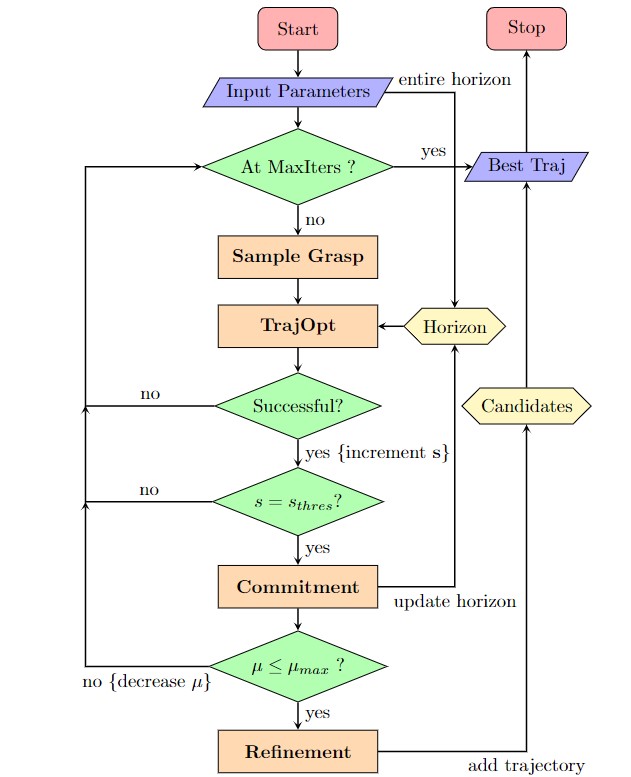}}
\caption{Flowchart representing the logic of the algorithm.}
\label{flowchart}
\end{figure}

The algorithm's parameters and variables are shown in Table \ref{tab:params_variables}, and a flowchart representation is shown in Fig. \ref{flowchart}. A summary of key aspects of the algorithm are as follows:

\begin{enumerate}

    \item \textbf{Initialization}: In the first iteration, a grasp pose, $\mathbf{G}$, is randomly sampled with respect to the target object and converted to the corresponding pose in the robot's task-space frame. Then, TrajOpt is executed with the initial parameters: $\mathbf{q}_0 = \mathbf{q}_s$, $N = N_0$, and $\mu = \mu_0$.
    
    \item \textbf{Commitment Step}: In any outer iteration, if TrajOpt successfully generates a solution trajectory, we store it in a temporary list of maximum size $s_{thres}$, and increment $s$ by 1. If $s=s_{thres}$, a commitment step is triggered, where we pick the lowest-cost trajectory from the temporary list and add a fraction of its waypoints to the committed list, $\mathcal{C}$. Then, we decrease the penalty coefficient, $\mu$ (unless it has reached $\mu_{thres}$), reduce $N$ by the number of newly-committed waypoints, increase $s_{thres}$, reset $s$ to 0, and empty the temporary list. If TrajOpt fails to find a solution, the algorithm proceeds to the next iteration without updating any variables.
    
    \item \textbf{Grasp Pose Sampling}: Using our parametrization model from Sec. \ref{sec:grasp_pose_parametrization}, a grasp pose is randomly sampled in each outer-loop iteration, and planning is performed for the current optimization horizon. This process continues until $\mathrm{MaxIters}$ is reached.
    
    \item \textbf{Upper Bound on Committed Waypoints}: To prevent an excessively long commitment list, an upper bound is imposed on the size of $\mathcal{C}$. Once reached, no further commitment steps take place. The fraction of waypoints added at each commitment step affects how quickly this bound is reached and is a hyper-parameter that requires tuning.
    
    \item \textbf{Refinement Step}: If TrajOpt successfully solves in any outer-loop iteration and the condition $\mu \leq \mu_{max}$ is satisfied, a refinement step is performed. In this step, the current solution trajectory is concatenated with the list of committed waypoints and passed to TrajOpt for further optimization. The resulting optimal trajectory spans the entire horizon and is added to a list of candidate solution trajectories.
    
    \item \textbf{Final Solution}: When $\mathrm{MaxIters}$ is reached, the final solution trajectory is selected from the candidate list based on the lowest cost. If the list is empty, indicating that no valid candidate trajectories were generated, the algorithm returns with failure. 
    
\end{enumerate}

\section{Experiments \& Results}

In this section, we present a comprehensive evaluation of our proposed approach by comparing it to several baseline methods. We conduct performance comparisons in four distinct task environments, which are pre-constructed and include simulations as well as real-world experiments using the Fetch robot. By systematically analyzing the results across these environments, we aim to assess the effectiveness and efficiency of our approach in different scenarios.

\begin{figure}[ht]
\centerline{\includegraphics[scale=0.54]{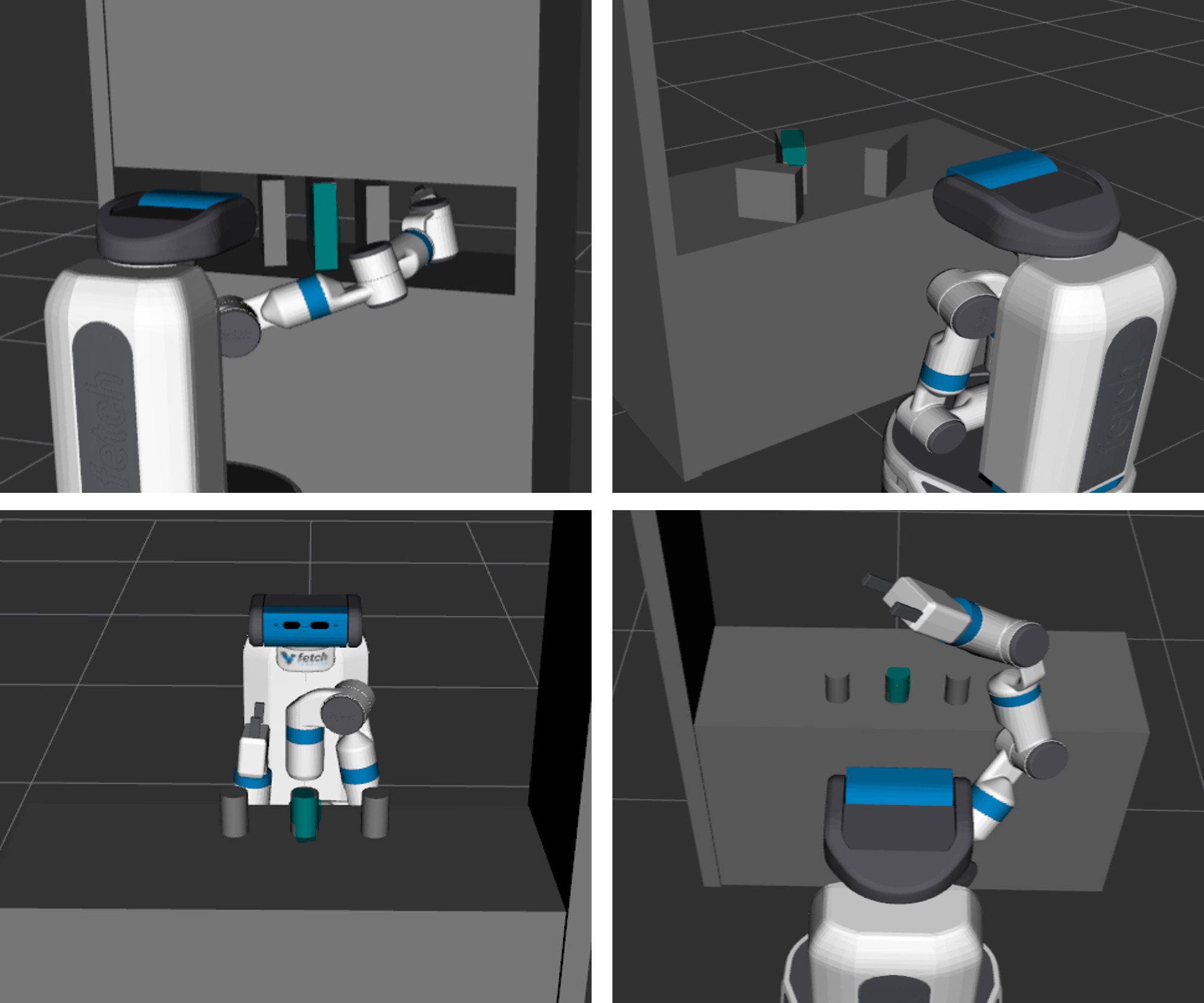}}
\caption{Visualisation of our algorithm planning a grasp towards a target (highlighted) in simulation, for 4 different scenes based on a shelf or tabletop workspace, and with multiple primitive objects in different poses. Each scene also plans from different initial arm configurations, including the constrained arm-tuck pose.}
\label{fig:sim4}
\end{figure}

\subsection{Baseline Approaches} \vspace{0.1cm}

For analysis of both the variable grasp pose planning and trajectory commitment ideas, we have designed the following 4 baseline approaches. Note that the baselines do not include any aspect of our trajectory commitment approach.

\begin{itemize}
    \item \textbf{RRT-Connect:} Randomly sample a grasp pose and run RRT-Connect once.
    
    \item \textbf{Fixed Goal:} Run the TrajOpt optimizer once, for a fixed hand-chosen grasp pose relative to the target object.

    \item \textbf{Variable Goal, Single Iteration:} Randomly sample a grasp pose relative to target object, and run TrajOpt once.

    \item \textbf{Variable Goal, Multi-Iteration:} Run TrajOpt 5 separate times, each with a different randomly sampled grasp pose, then return trajectory with lowest cost.

\end{itemize}

\subsection{Experimental Scenes} \vspace{0.1cm}

For experimentation, we have constructed 4 different task scenes. They are mostly distinct in the initial arm configuration, the objects' shapes, dimensions and poses, and whether the workspace is a shelf-like or table-top scenario. Figure \ref{fig:sim4} shows the following scenes, \textbf{Box Shelf}, \textbf{Box Table}, \textbf{Cylinder 1} and \textbf{Cylinder 2}, in the order of left to right, top to bottom. The target object in each scene is highlighted.

\begin{itemize}
    \item \textbf{Box Shelf} - A shelf-like scene, with 3 large boxes (25cm $\times$ 25cm $\times$ 8cm) arranged in a row. Initially the arm is inside the shelf.

    \item \textbf{Box Table} - A table-top scene, with 3 medium-sized boxes (15cm $\times$ 15cm $\times$ 6cm) arranged such that the target is occluded by another collision object. Initially in arm-tuck position.

    \item \textbf{Cylinder 1} - A table-top scene, with 3 small cylinders arranged in a row. Initially in arm-tuck position.

    \item \textbf{Cylinder 2} - Same as the \textbf{Cylinder 1} scene, apart from the arm initially being in one of our randomly-generated configurations, which from previous experimentation is a very easy pose to plan from.
\end{itemize}


\subsection{Metrics}

For both the simulated and real world experiments, we base our discussion on the following three metrics: 

\begin{itemize}
    \item Planning Time (s)
    \item Trajectory Cost (using the cost definition from TrajOpt)
    \item Success Rate (\%)
\end{itemize}

\subsection{Simulation Results}

We ran 300 independent trials for each of the 5 approaches (4 baselines + Ours), for each of the 4 different task scenes shown in Fig \ref{fig:sim4}. The results are shown in Table \ref{table:simulation_results}.

\begin{table*}[htbp]
\centering
\begin{tabular}{|l|ccc|ccc|ccc|ccc|}
\hline
& \multicolumn{3}{c|}{Box Shelf} & \multicolumn{3}{c|}{Box Table} & \multicolumn{3}{c|}{Cylinder 1} & \multicolumn{3}{c|}{Cylinder 2}\\ \hline

& $\%$ & $-$ & sec & $\%$ & $-$ & sec & $\%$ & $-$ & sec & $\%$ & $-$ & sec \\ \hline

\rowcolor{LightCyan}
Method & Succ. & Cost & Time & Succ. & Cost & Time & Succ. & Cost & Time & Succ. & Cost & Time \\ \hline

RRT-Connect\cite{RRTconnect}  & 22     & 0.540 & \textbf{4.3} & 28 & 1.99 & \textbf{1.3} & 36 & 1.77 & \textbf{1.3} & 37 & 9.76 & \textbf{1.3} \\ \hline

\rowcolor{LightGray}
Fixed Goal & -     & 0.130 & 12.9 & - & 0.69 & 25.6 & - & 0.72 & 20.6 & - & 4.81 & 17.3 \\ \hline

Variable Goal, Single Iteration & 19     & 0.130 & 15.6 & 20 & 0.69 & 30.8 & 33 & 0.72 & 22.0 & 34 & 4.83 & 16.4 \\ \hline

\rowcolor{LightGray}
Variable Goal, Multi-Iteration & 82     & 0.128 & 77.8 & 74 & 0.7 & 158.5 & 93 & 0.71 & 211.5 & \textbf{100} & \textbf{4.74} & 81.8 \\ \hline

Variable Goal, Multi-Iteration, Commitment (Ours)            & \textbf{95}     & \textbf{0.122} & 79.8 & \textbf{97} & \textbf{0.61} & 123.3 & \textbf{96} & \textbf{0.66} & 208.2 & \textbf{100} & 4.88 & 96.9 \\ \hline

\end{tabular}
\caption{Simulation results from 300 trials for each scenario per method. Best results are displayed in \textbf{bold}.}
\label{table:simulation_results}
\vspace{-0.5cm}
\end{table*}

In the simulated scenes, our approach by far had the highest success rate in all scenarios. Our approach also consistently generated the lowest-cost trajectories compared to the baselines in three out of four scenarios. For the \textbf{Cylinder 2} scene, where the initial arm configuration is unconstrained, the \textbf{Variable Goal, Multi-Iteration} baseline generated slightly lower-cost trajectories on average, while matching the 100$\%$ success rate of our approach. For the \textbf{Fixed Goal} baseline, we chose to not include its success rate, as the fixed grasp poses it plans towards are hand-picked for each scene to guarantee feasibility. Since all 4 other approaches adopt the automated random grasp sampling, the success rate is not a meaningful metric for comparison here.

Regarding computing time, the \textbf{RRT-Connect} baseline consistently exhibited the shortest computation time for generating a trajectory. However, it's important to note that the trajectories generated by this baseline were significantly less efficient in terms of cost compared to trajectory optimization methods. It's also worth noting that both our approach and the \textbf{Variable Goal, Multi-Iteration} baseline exhibited significantly longer planning times compared to the single-iteration trajectory optimization baselines, as a consequence of running TrajOpt for multiple grasp poses.

Overall, the results demonstrate that our approach outperforms the baselines in terms of generating lower-cost trajectories and achieving higher success rates, especially for constrained tasks involving cluttered objects and challenging initial configurations. However, the improvements were less significant in less-constrained task environments.

During the simulated experiments, we made an observation that confirmed one of our previous hypotheses. As our approach committed more trajectory waypoints, resulting in a shrinking optimization horizon, TrajOpt's speed in generating a solution trajectory or returning failure significantly increased. This speed-up is attributed to the fact that the dimension of the optimization problem is proportional to the number of trajectory waypoints to plan for. Consequently, our approach was able to plan for approximately double the number of outer-loop iterations (the number of times the TrajOpt optimizer is run) compared to the \textbf{Variable Goal, Multi-Iteration} baseline, within a similar total planning time. This improvement directly contributed to the higher success rate achieved by our approach.

\subsection{Real Robot Results}

\vspace{-0.3cm}
\begin{figure}[ht]
\centerline{\includegraphics[scale=0.26]{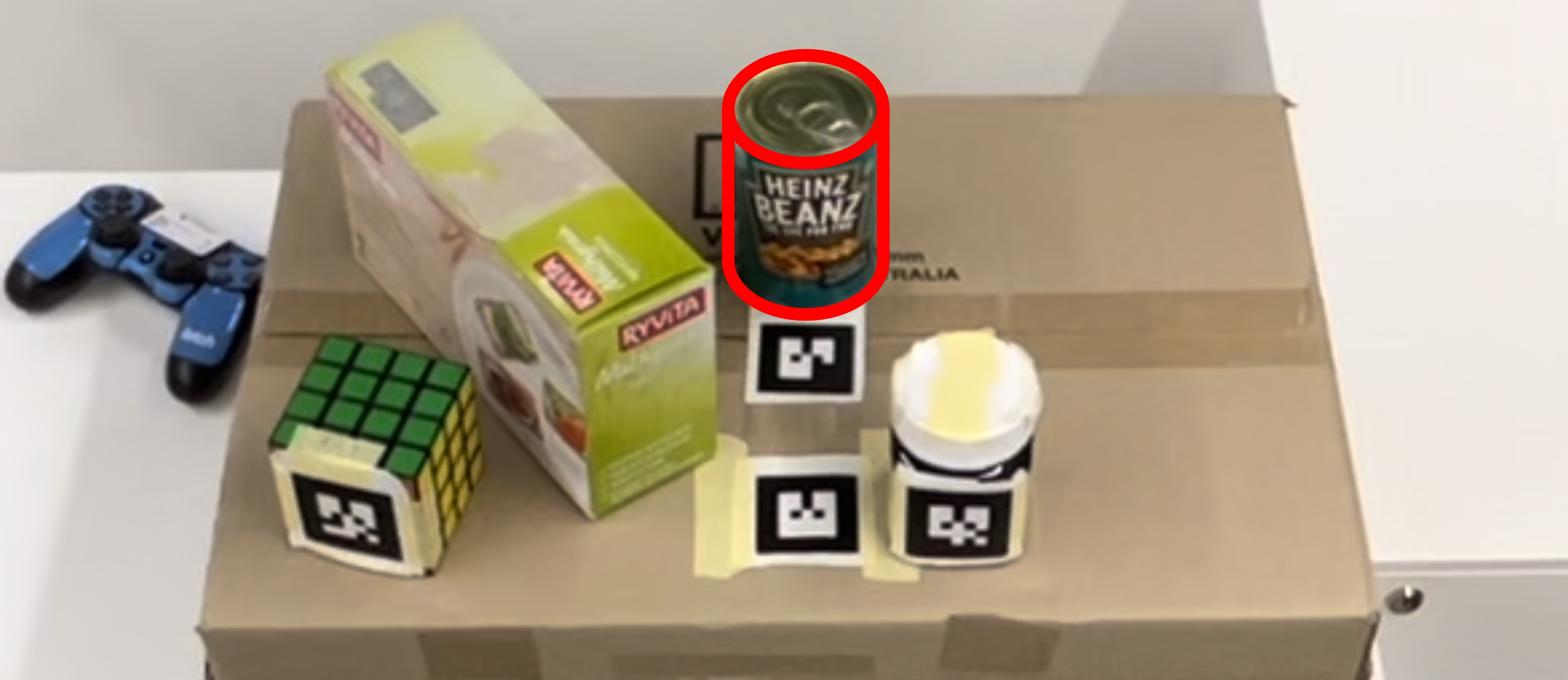}}
\caption{Top view of our real-life experimental scene. The cylinder (highlighted) is fixed to be the target object across all trials.}
\label{fig:real_scene}
\end{figure}

In our real-world experiments, we employed the Fetch robot within a table-top environment featuring a fixed cylindrical target object and several cuboids of varying dimensions and poses (see Fig. \ref{fig:real_scene}). To simplify the pose determination of all objects, including the table, we utilized Alvar markers. We conducted 10 independent trials for each of the three approaches, and the results are shown in Table \ref{table:real_results}.

The real-world experiment results demonstrate that our approach consistently generated lower-cost trajectories compared to the other two baselines. Moreover, both our approach and the \textbf{Variable Goal, Multi-Iteration} baseline exhibited a remarkable improvement in success rate, succeeding in 8 out of the 10 trials. This significant enhancement in robustness is accompanied by a trade-off of increased planning time. However, it is important to note that the planning times for our approach and the \textbf{Variable Goal, Multi-Iteration} baseline were comparable. 

\begin{figure*}[t]
\centering
\includegraphics[scale=0.956]{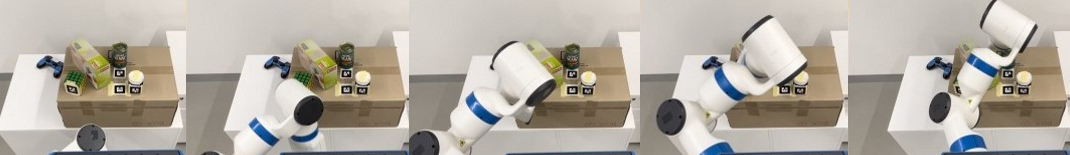}\\
 \vspace{0.05cm}
\includegraphics[scale=0.6305]{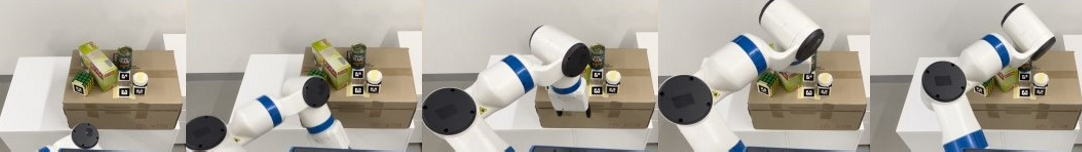}
\caption{Trajectories generated by the \textbf{Variable Goal, Multi-Iteration} baseline (top-row) and our approach (bottom-row). Our approach produced a lower-cost trajectory.}
\label{fig:goodmed_compare}
\vspace{-0.8cm}
\end{figure*}

For a visual comparison of the solution trajectories generated by our approach and the \textbf{Variable Goal, Multi-Iteration} baseline, Figure \ref{fig:goodmed_compare} illustrates the qualitative difference between the solutions of the two approaches, highlighting the superior performance of our approach in terms of trajectory quality.

\begin{table}[t!]
\centering
\begin{tabular}{|l|ccc|ccc|ccc|ccc|}
\hline

& $\%$ & $-$ & sec \\ \hline

\rowcolor{LightCyan}
Method & Succ. & Cost & Time \\ \hline

Variable Goal, Single Iteration & 40     & 0.76 & \textbf{23.6} \\ \hline

\rowcolor{LightGray}
Variable Goal, Multi-Iteration  & \textbf{80}     & 0.68 & 80.7 \\ \hline

Variable Goal, Multi-Iteration, Commitment (Ours)             & \textbf{80}     & \textbf{0.42} & 80.1 \\ \hline

\end{tabular}
\caption{Real robot results from 10 trials for each method. Best results are displayed in \textbf{bold}.}
\label{table:real_results}
\end{table}

\section{Discussion}

Of the four simulation scenes we designed, our algorithm consistently generated the lowest-cost trajectories and exhibited the highest success rate in three of them. However, in the less constrained environment of the \textbf{Cylinder 2} scene, the \textbf{Variable Goal, Multi-Iteration} baseline outperformed our approach in trajectory cost while maintaining a similar level of consistency. Consequently, we hypothesize that our approach may not offer a significant advantage over existing motion planners in easier task environments. However, it excels in highly constrained tasks by efficiently selecting better grasp poses and consistently generating corresponding lower-cost trajectories.

Regarding our experimental design, a key limitation is the lack of variation in the task scenes. The \textbf{Box Table} and \textbf{Cylinder 1} scenes feature a table-top environment with three objects of the same type scattered on the table, while the \textbf{Box Shelf} and \textbf{Cylinder 2} scenes have similar arrangements. Throughout the 300 simulation trials conducted for each scene, all aspects of the environments remained unchanged. This design choice enables consistent calculation of success rates for each approach. Additionally, since most of the approaches, including our method, incorporate random sampling, conducting repeated trials in consistent task scenes allows for more accurate interpretation of performance metrics using mean and standard deviation values.

However, this design also limits the ability to draw broader conclusions about the performance of our method. To achieve more generalizable conclusions, it may be necessary to introduce higher variation in experimental scenes by incorporating different collision object arrangements across trials. One approach is to randomly generate a large number of distinct, highly constrained task scenes and conduct slightly fewer repeated trials for each scene. This method provides the distribution of outcomes for each scene while still utilizing mean and standard deviation as valid metrics to represent the performance of each approach. Moreover, this approach introduces greater scene variation, which helps reduce systematic bias towards scenes with specific characteristics.

Another potential improvement in scene design involves the choice of initial arm configurations. Generating random initial configurations that avoid immediate collisions with objects in the highly constrained task space can be challenging. One solution is to position the robot directly in a feasible pose within the task environment and then randomly sample a large number of task space goal poses for the end-effector. An efficient trajectory planning algorithm such as RRT-Connect can then be used to move the arm towards each task space goal pose. Successful trajectories from this process yield valid joint-space arm configurations that can be used as initializations. Overall, these adjustments to our experimental design could lead to more powerful conclusions regarding the performance and robustness of our method, which would generalize well to a wider range of scenes.


\section{Conclusions}

In this paper, we introduced the concept of variable grasp pose planning and trajectory commitment, and the key motivations behind these ideas. Through both simulated and real experiments, we demonstrated that our proposed algorithm exhibits more consistent and robust performance across a series of four task scenes compared to various baseline approaches based on existing motion planning algorithms. However, we also discussed the limitations of our experimental design, which provides valuable insights for future improvements.

One immediate focus for future work is the design of a method to generate randomized task scenes that incorporate greater variation in object types, dimensions, positions, orientations relative to the robot, and the initial configuration of the robot itself. By testing the proposed method in a broader range of settings, we can obtain more meaningful results and evaluate its performance under diverse conditions. Additionally, improving the computation time for the proposed method, such as by parallelizing computing trajectories of multiple grasp poses \cite{murooka}, will allow our method to be even more competitive with existing approaches.

For robotic manipulators with a mobile base, such as the Fetch robot, an intriguing aspect to explore is incorporating the base pose of the robot as additional degrees of freedom in grasp planning and trajectory optimization. Preliminary results suggest that the base pose can directly impact the optimal grasp pose and trajectory cost for moving the arm manipulator toward the target object. Therefore, investigating the influence of the base pose on the overall performance and cost of the algorithm holds promise for further advancements.

In summary, this paper presents a novel approach for variable grasp pose planning and trajectory commitment, showcasing its superiority over baseline approaches in terms of consistency and robustness. While acknowledging the limitations of our experimental design, we propose future directions that can build upon the concepts and methods introduced here. These future endeavors have the potential to address current limitations, enhance the overall performance, and increase the robustness of the algorithm in real-world scenarios.


\bibliographystyle{unsrt}
\bibliography{trajopt}

\end{document}